\pdfoutput=1
\documentclass[11pt]{article}

\usepackage[]{EMNLP2022}

\usepackage{times}
\usepackage{latexsym}
\usepackage{graphicx}
\usepackage{float}
\usepackage{multirow}
\usepackage{multicol}
\usepackage{booktabs}
\usepackage{enumerate}
\usepackage{verbatim}
\usepackage{colortbl}

\usepackage{mydef}

\usepackage[T1]{fontenc}
\usepackage[utf8]{inputenc}

\usepackage{microtype}

\title{$\mathcal{Y}$-Tuning: An Efficient Tuning Paradigm for Large-Scale Pre-Trained Models via Label Representation Learning}

\author{Yitao Liu \and Chenxin An \and Xipeng Qiu \\
  Fudan University \\
  \texttt{\{yitaoliu20, cxan20, xpqiu\}@fudan.edu.cn}}

\begin{document}
\maketitle
\begin{abstract}
With the success of large-scale pre-trained models (PTMs), how efficiently adapting PTMs to downstream tasks has attracted tremendous attention, especially for PTMs with billions of parameters. Previous work focuses on designing  parameter-efficient tuning paradigms but needs to save and compute the gradient of the whole computational graph. In this paper, we propose $\mathcal{Y}$-Tuning, an efficient yet effective paradigm to adapt frozen large-scale PTMs to specific downstream tasks. $\mathcal{Y}$-tuning learns dense representations for labels $\mathcal{Y}$ defined in a given task and aligns them to fixed feature representation. Without computing the gradients of  text encoder at training phrase, $\mathcal{Y}$-tuning is not only parameter-efficient but also training-efficient. Experimental results show that for $\text{DeBERTa}_\text{XXL}$ with 1.6 billion parameters, $\mathcal{Y}$-tuning achieves performance more than $96\%$ of full fine-tuning on GLUE Benchmark with only $2\%$ tunable parameters and much fewer training costs. 

\end{abstract}

\section{Introduction}

Large-scale pre-trained language models (PTMs) can capture general language knowledge from a large corpus and have become backbone models for many NLP tasks~\cite{Devlin19bert,openai20gpt3,qiu20ptm}.
However, how effectively adapting their knowledge to downstream tasks is still a key problem.

Currently, the prevalent paradigm of adapting PTMs to downstream NLP tasks is fine-tuning. While fine-tuning obtains good performance, it is parameter-inefficient and training-inefficient. A natural solution of this problem is \textit{lightweight fine-tuning}, which freezes all (or most) of the pre-trained parameters and augments the model with small trainable modules. There are two main paradigms of lightweight fine-tuning. (1) Adapter-tuning \cite{houlsby2019parameter,stickland2019bert}, which inserts additional task-specific adaption modules into PTMs, achieving comparable performance with fine-tuning while adding only around 2-4\% task-specific parameters. (2) Prompt (or Prefix) tuning \cite{Li2021Prefix,Lester2021Power,liu2021gpt}, which fixes the parameters of PTMs and modifies the model behavior by adding some learnable prompt vectors as the prefix of input sequences. These prompts (or prefixes) can affect how subsequent input is processed.

However, adapter-tuning and prompt-tuning just improve the parameter-efficiency rather than training-efficiency. In the training phase, even with few tunable parameters, they still need to save the computational graph for gradient descent resulting in huge training costs. We argue that training-efficiency is also crucial in real application scenarios. On the one hand, models are often required to be capable of rapid iteration in  industry. Spending too much time on training leads to a significant increase in training costs and prolongs the iteration cycle. On the other hand, currently very-large-scale PTMs such as GPT-3~\cite{openai20gpt3} and ERNIE 3.0~\cite{ernie3.0} are deployed on cloud servers and only \textbf{forward} APIs are accessible~\cite{sun2022bbt}. Users need to send texts as queries to the servers without the permission of updating the PTM which hinders the application of many parameter-efficiency methods such as adapter-tuning~\cite{houlsby2019parameter}.  

In order to find a cheaper and more efficient way to train large-scale PTMs, an intuitive idea is to tune additional modules appropriately while no backpropagation is required on large models. In this paper, we propose $\mathcal{Y}$-Tuning, a new paradigm to adapt Transfomer-style large PTMs to downstream NLP tasks. Similar to prompt tuning, $\mathcal{Y}$-Tuning also seeks to make use of the semantic information of labels. Instead of building verbalizers to bridge labels and words, and then incorporating label information into natural language prompt, we could choose to learn dense semantic label representations and align them to fixed feature representations of input text.
We assume that PTMs can capture the generic language and world knowledge implied in large-scale training data. The task-specific features can be induced from the frozen generic representation by powerful tuning on the label side. 

The primary contribution of $\mathcal{Y}$-tuning is to learn a more powerful label representation close to the frozen feature representation, instead of tuning the feature representation to adapt the labels.
The advantages of $\mathcal{Y}$-Tuning can be summarized as follows:
\begin{itemize*}

    \item \textbf{Parameter-Efficiency.} $\mathcal{Y}$-tuning
    freezes text encoder and learns label representations with a lightweight model  for downstream tasks. For large PTMs such as DeBERTa, $\mathcal{Y}$-tuning only tunes $2\%$ task-specific parameters compared with fine-tuning. (Cf. Sec~\ref{sec: quatitative results})

    \item \textbf{Training-Efficiency.} $\mathcal{Y}$-tuning does not need gradients of large PTMs in the training phase leading to less computing memory and training time  for utilizing large PTMs in practice. Sec.~\ref{sec:training efficiency} shows that $\mathcal{Y}$-tuning is more than $6$ times faster than that of fine-tuning and other lightweight tuning methods with less GPU memory.

    \item \textbf{Model Robustness.} $\mathcal{Y}$-tuning does not tune the feature representation of the input sequence, which makes the $\mathcal{Y}$-tuned models more robust and difficult to text-based attack. (Cf. Sec~\ref{sec:robustness})

\end{itemize*}

\newcommand{\rulesep}{\unskip \vrule width 0.5pt}

\begin{figure*}[t]
  \centering
\subfloat[Fine-Tuning]{\label{fig:cnn}
  \includegraphics[scale=0.6]{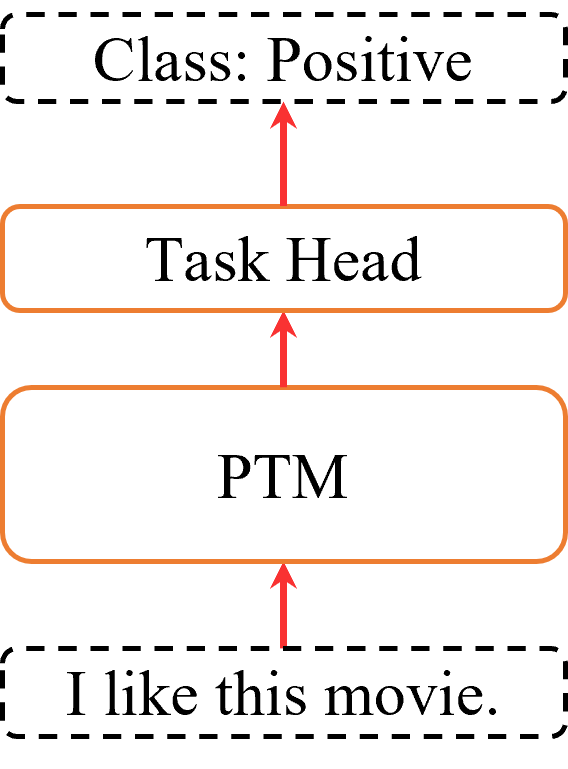}
}
\rulesep
\subfloat[Adapter-Tuning]{\label{fig:cnn}
  \includegraphics[scale=0.6]{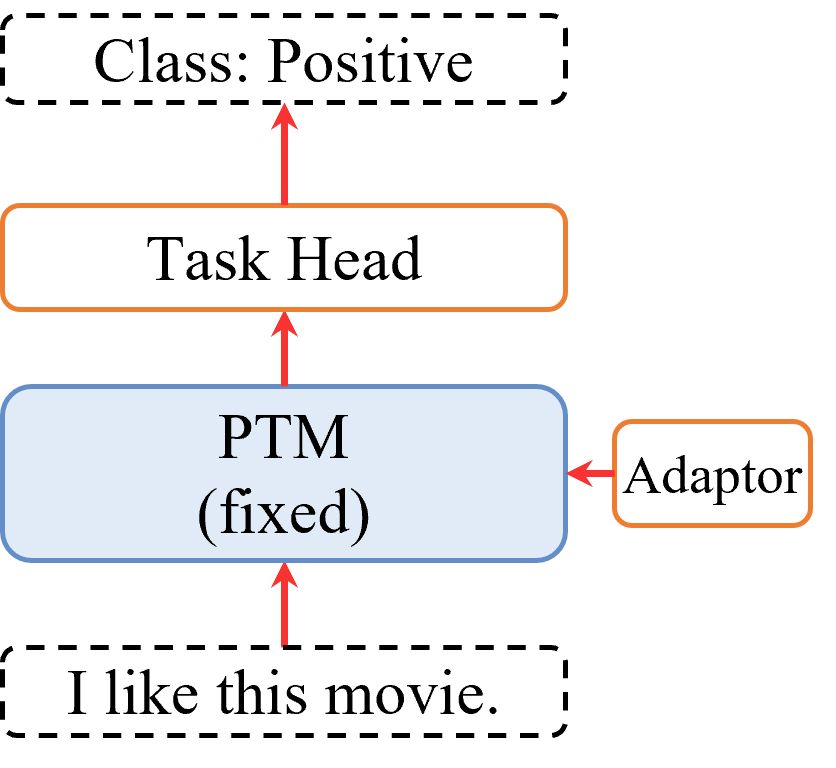}
}
\rulesep
%\hspace{2em}
\subfloat[Prompt-Tuning]{\label{fig:fig-prompt}
  \includegraphics[scale=0.6]{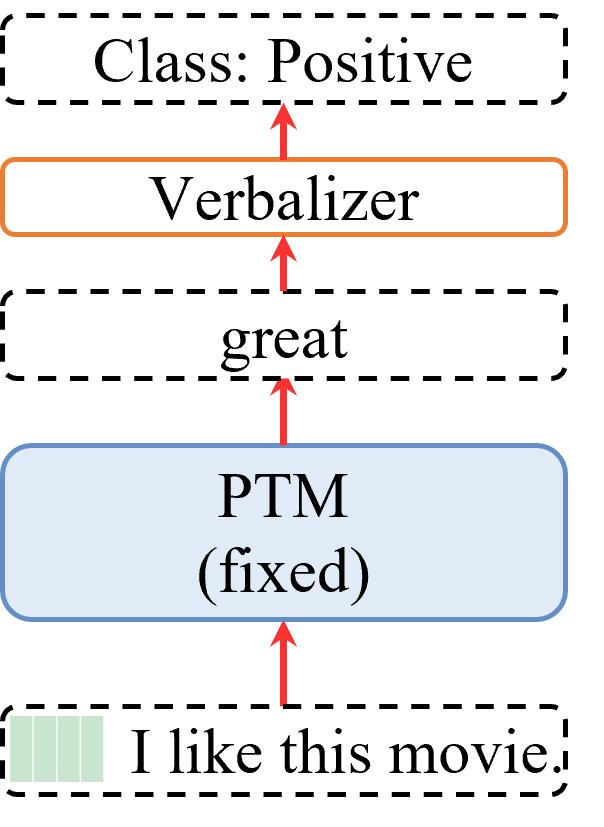}
}
\rulesep%\hspace{2em}
\subfloat[$\mathcal{Y}$-Tuning]{\label{fig:fig-ytuning}
  \includegraphics[scale=0.27]{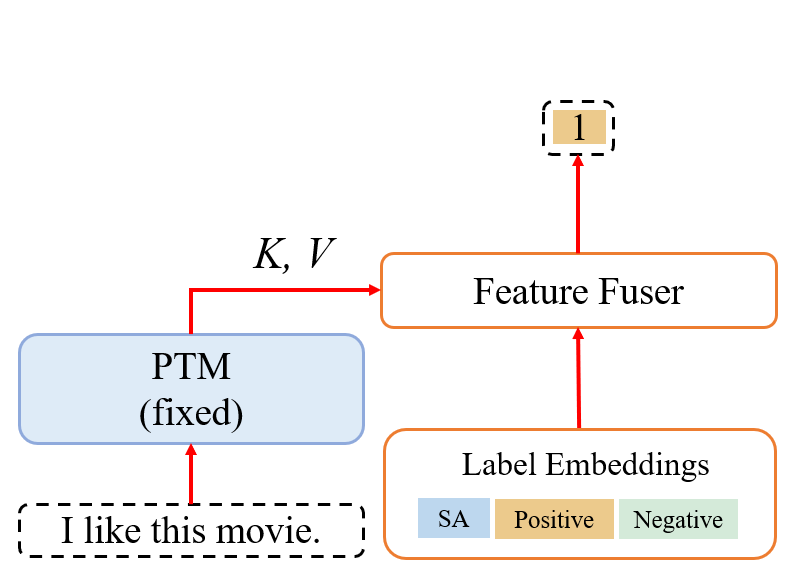}
}
  \caption{Illustration of Four Tuning Paradigms. In (c), \includegraphics[scale=0.5]{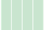} represents continuous prompt vectors.}
  \label{fig:tuning-paradigms}
\end{figure*}

\begin{table*}
\centering
\begin{tabular}{lcccccc}
\toprule
Tuning Type & Input & Output & Function & \tabincell{c}{Tunable\\ Modules} & \tabincell{c}{Param.\\ Efficiency} & \tabincell{c}{Training\\ Efficiency}\\
\midrule
Fine-Tuning & $x$ & $p(y|x)$ & $\red{f}\circ\red{\phi(x)}$ & $\red{f},\red{\phi}$ & \xmark & \xmark\\
Feature-based-Tuning & $x$ & $p(y|x)$ & $\red{f}\circ\cyan{\phi^*}(x)$ & $\red{f}$& \cmark & \cmark\\
Adapter-Tuning & $x$ & $p(y|x)$ & $\red{f}\circ\cyan{\phi^{*+\red{\delta}}}(x)$ & $\red{f},\red{\delta}$& \cmark & \xmark\\
Prompt-Tuning & $x$ & $p(y|x)$ & $\red{f}\circ\cyan{\phi^*}([\red{p};x])$ & $\red{f}$,$\red{p}$& \cmark & \xmark\\
$\mathcal{Y}$-Tuning & $x,\mathcal{Y}$ & $p(c|x,\mathcal{Y})$ & $\red{f}(\red{\psi}(\mathcal{Y}),\cyan{\phi^*}(x))$ & $\red{f},\red{\psi}$& \cmark & \cmark\\
\bottomrule
\end{tabular}
\caption{The tuning paradigms of PTMs.
$x$ is a series of tokens and $y$ is a single class label. $\phi$ is pre-trained model for feature representation and $f$ can be regarded as an task-specific function.
$f\circ\phi$ denotes a composite function and $\phi^*$ indicates the parameters are frozen.
In adapter-tuning, the  feature function $\phi^{*+\red{\delta}}$ contains some extra tunable parameters $\red{\delta}$. In prompt-tuning,
$p$ consists of additional tunable tokens per downstream task to be prepended to the input text. In $\mathcal{Y}$-tuning, $\mathcal{Y}$ denotes the label set, $c$ denotes the indexes of label and $\psi$ denotes the label embeddings function.
}
\label{tb:paradigmshift}
\end{table*}

\section{Preliminaries of Tuning PTMs}

An NLP task usually takes an input text $x\in \mathcal{X}$ and predicts a label $y\in \mathcal{Y}$, where $\mathcal{X}$ is discrete feature space of natural language and $\mathcal{Y}$ is label space. Let $\Phi=\{\phi: \mathcal{X}\rightarrow \mathcal{Z} \}$ denote a collection of functions mapping the original language space $\mathcal{X}$ to some latent semantic feature space $\mathcal{Z}$, and $\mathcal{F}=\{f: \mathcal{Z}\rightarrow \mathcal{Y} \}$ be a collection of task-specific classification functions from the feature space to the label space $\mathcal{Y}$.
Give a training set $\mathcal{D}$ for a specific NLP task, we define a loss function $\ell: \mathcal{Y}\times \mathcal{Y} \rightarrow \mathbb{R}^+$ and find the optimal representation and classification functions by
\begin{align}
\phi^*, f^* = \mathop{\arg\min}_{f\in \mathcal{F},\phi\in{\Phi}} \sum_{(x,y)\in \mathcal{D}} \ell\big(f\circ\phi(x),y\big),
\end{align}
where $f\circ\phi$ denotes a composite function.

\subsection{Fine-Tuning}
In fine-tuning, we initialize the representation function $\phi$ with a PTM and optimize its parameters with a small amount on the specific downstream task. Although fine-tuning is the currently most effective paradigm to utilize PTMs on downstream tasks, its main drawback is its parameter-inefficiency.

\subsection{Adapter-Tuning}
Recently, a lightweight tuning solution is to inject small fine-tunable adaptation modules into PTMs while freezing the original parameters of PTM. \citet{stickland2019bert} equipped a single share BERT model with small additional task-specific adaptation modules, and projected attention layers (PALs).
Similarly, \citet{houlsby2019parameter} modified the architecture of pre-trained BERT by adding adapter modules.

In adapter-tuning, the PTM is shared and the adaption modules are separately fine-tuned on different tasks. Although adapter-tuning is parameter-efficient and has roughly 5-10 times fewer tunable parameters, it still needs to keep the whole computation graph of PTM in the training phase. Thus, its computational cost is still expensive. The main benefit of adapter-tuning claimed by ~\citet{pfeiffer2020adapterhub} is saving storage space.
%The parameters of the original network remain fixed, yielding a high degree of parameter sharing.

\subsection{Prompt-Tuning}

Another lightweight tuning approach is reformulating the downstream tasks into an MLM task by designing appropriate prompts. Prompt-based adaptation have shown great power in few-shot setting~\cite{openai20gpt3,Scao2021How,Schick2021Exploting}, and zero-shot setting~\cite{petroni2019language,Jiang2020How}.
Unfortunately, the effectiveness of prompt-based tuning depends on the quality of prompt which usually requires human involvement. As a result, the performance of prompt-based adaptation still lags far behind fine-tuning on many downstream tasks.

Recently, prompt tuning\cite{Li2021Prefix,Lester2021Power,liu2021gpt} makes the prompt-based tuning a very promising method for an efficient serving of large-scale PTMs, which inserts continuous prompt as the prefix of the input,
Prompt tuning as a parameter-efficient tuning technique achieved comparable performance in fully-supervised setting and outperformed model fine-tuning in few-shot setting. However, prompt-tuning is also training-inefficient. It cannot avoid both computing memory and time cost for adapting PTMs to downstream tasks.

\section{$\mathcal{Y}$-Tuning}
\label{sec:Y-tuning-sec}
we consider that the gap between pre-training and fine-tuning can be narrowed from both sides: the feature side and label side. If we cannot tune the feature side due to resource constraints, tuning the label side is also an effective method to boost the performance of PTMs on downstream tasks. We show illustrations of the paradigm of $\mathcal{Y}$-Tuning compared with other tuning paradigms in Figure ~\ref{fig:tuning-paradigms}. Table \ref{tb:paradigmshift} 
formally exhibits different tuning paradigms and compares them on efficiency dimension. In the following parts, we will give a detailed description of our proposed method.

\subsection{Overall Method}

Instead of tuning feature representation $\phi^*(x)$ to the task-specific label space, we fix the feature representation and learn a task-specific label representation $\psi(\mathcal{Y})$.
Then we use a fuse and score function $f$ to induce the relevant input feature to label representation and output a score for each label indicating the possibility to be the correct label.
The loss function is triplet loss, which maximizes the score of the correct label $y$ while minimizing the score of the wrong label $y'$. The triplet loss is defined as
\begin{equation}
\small
\label{eq:loss}
\mathcal{L}_{\psi,f} =   \sum_{y'\in \mathcal{Y}}\Big[f\big(\phi^*(x),\psi(y')\big)
 -f\big(\phi^*(x),\psi(y)\big) + \gamma\Big]_{+},
% \mathcal{L}_{\psi,f}(x,y)= \sum_{\overset{y'\in \mathcal{Y}}{  y'\neq y}} \Big[f\big(\phi^*(x),\psi(y')\big)\hspace{1.2cm} \nonumber\\
%  -f\big(\phi^*(x),\psi(y)\big) + \alpha\Big]_{+},
\end{equation}
where $[\cdot]_+=\max(\cdot,0)$ and $\gamma$ is the margin between positive and negative pairs.
% Give a training set $\mathcal{D}$ for a specific NLP task, we find the optimal representation and function by
% \begin{align}
% \psi^*, f^* = \mathop{\arg\min}_{f\in \mathcal{F},\psi\in{\Psi}} \sum_{(x,y)\in \mathcal{D}} \mathcal{L}_{\psi,f}(x,y).
% \end{align}

In inference phase, we make the prediction by
 \begin{align}
\hat{y}=\mathop{\arg\max}_{y\in \mathcal{Y}} f^*\left(\phi^*(x),\psi^*(y)\right),
\end{align}
where $\phi^*$ is frozen pre-trained model, $f^*$ and $\psi^*$ are well  learned functions by $\mathcal{Y}$-Tuning.

\subsection{Architecture Designing}

We build the architecture from three components:
\begin{enumerate*}
  \item A frozen pre-trained {feature encoder} $\phi^*$ that encodes input texts into generic features.
   \item A label embedding function  $\psi$ that represents the set of labels in the embedding space.
  \item A {label-aware feature fuser} $f$ that fuses semantic features into label representations. The most possible label is chosen from these feature-enhanced label representations.
\end{enumerate*}

\paragraph{Frozen Feature Encoder}
Feature encoders $\phi^*$ are large-scale and powerful PTMs~\cite{Devlin19bert, ernie3.0} which project the natural language input $x$ into meaningful vectors $\phi^*(x)$ in the embedding space. In $\mathcal{Y}$-tuning we use $\phi^*$ with parameters frozen.

\paragraph{Label Embeddings}
Given a label set $\mathcal{Y}$ comprised of $N$ labels, we map label $y_{c}\in\mathcal{Y} $ to one or several continuous vectors $\psi(y_{c})$ for each label index $c\in\{1, \cdots, N\}$. We also introduce a task representation to select task-specific information from input features:
\begin{align}
\psi(\mathcal{Y})=[\be_T;\be_{1};\cdots;\be_{c};\cdots] \in {\mathbb{R}^{N\times D}}, \label{eq:label representation}
\end{align}
where $\be_T$ denotes the task embedding and $\be_c$ denotes the embedding for $c$-th label; $D$ denotes the dimension of label embeddings. Each label can be represented by multiple embedding vectors and there are several ways to initialize label embeddings. See Appendix \ref{sec:ablations} for ablation results.

\paragraph{Label-Aware Feature Fuser}

Now we need to enable label embeddings $\psi(\mathcal{Y})$ to obtain relevant semantic information from text features $\phi^*(x)$ and then obtain scalar scores for each label through semantic similarity calculation. To achieve this we introduce a label-aware feature fuser.

The feature fuser is a task specific function $f$ that fuses $\phi^*(x)$ and $\psi(\mathcal{Y})$ and scores labels. We implement the fusing module by cross attention mechanism using a Transformer decoder layer. The difference is that all attention modules here are non-causal and we also use the full self-attention to introduce interactions between labels. After label and task representations are fully integrated with text information, we score each label as the cosine similarity between task feature and its own feature.

With label scores $f(\phi^*(x), \psi(\mathcal{Y}))$, we use triplet loss function described in Eq.~\ref{eq:loss} as object function in training phase. And in inference phase the label with maximum score is chosen to be model's prediction.

\subsection{Training Efficiency Analysis}
Assuming that we have a $L$ layers PTM with Transformer blocks, the complexity\footnote{Here we mainly compute of the complexity of QKV attention.} of PTM is $\mathcal{O}(LM^2)$, where $M$ is the input sequence length.

The prompt-tuning pre-pends $P$ continuous prompts to input sequence, and the complexity becomes $\mathcal{O}(L(M+P)^2)$. The complexity increases greatly even for a small $P$. When the PTM is very large, the cost of prompt-tuning is unaffordable, especially for the training phase.

In $\mathcal{Y}$-tuning, the complexity of self-attention and cross-attention in label perceiver are $\mathcal{O}(N^2)$ and $\mathcal{O}(MN)$ respectively, where $N$ be the size of label set. Since $N\ll M$ for most downstream tasks, especially for classification, the increased computational cost is negligible compared to the cost of the original PTM. Compared to prompt-tuning, $\mathcal{Y}$-tuning has a lower computational cost in both training and inference phases.

\section{Generalized $\mathcal{Y}$-tuning}

Labels in Sec~\ref{sec:Y-tuning-sec} are sentence-level and mainly suitable for text-classification tasks. Directly extending $\mathcal{Y}$-tuning to other types of tasks such as sequence labeling is not trivial since there would be $NM$ labels if we maintain relative labels for each token.

\begin{figure}[h]
  \centering
  \includegraphics[scale=0.3]{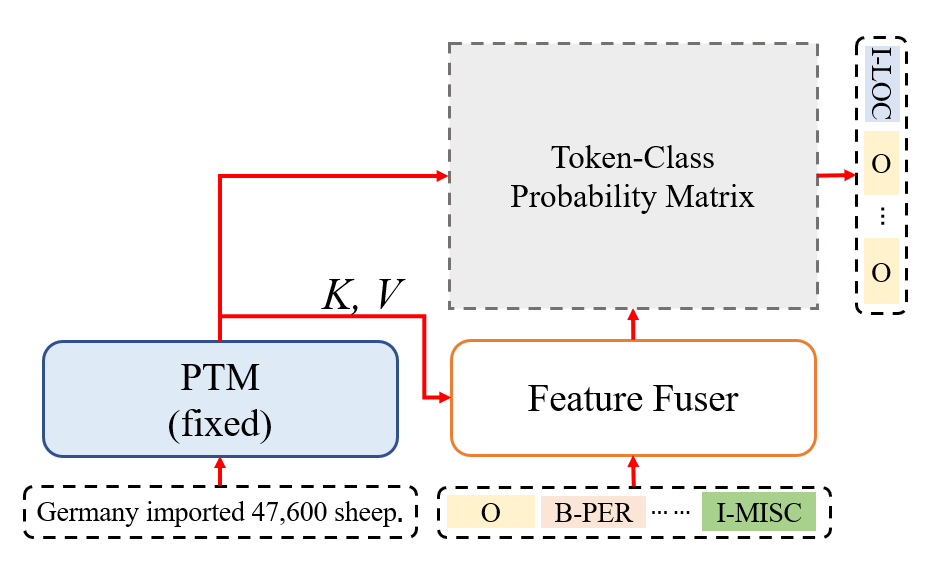}
  \caption{The architecture of Generalized $\mathcal{Y}$-Tuning for sequence labeling task.}
  \label{fig:fig-y-tuning-seqlab-arch}
\end{figure}

For sequence labeling, an alternative way is still applying $N$ labels corresponding to token tags such as 
\texttt{"B-PER"} in named entity recognition (NER). We keep the label-aware feature fuser the same as described in Sec~\ref{sec:Y-tuning-sec} except for the task representation which is unused and we take the dot-product of label representations and PTMs' output hidden states. After normalizing by softmax along the token dimension, we get the token-label probability matrix. For each token's probability distribution, we make the prediction by argmax. Cross entropy is chosen as the loss function for training. The architecture is shown in fig~\ref{fig:fig-y-tuning-seqlab-arch}. Such method can be regarded as a tag-based dynamic clustering.

Generalizing $\mathcal{Y}$-tuning to span-based question answering task is quite similar. We select \texttt{"BEGIN"} and \texttt{"END"} as labels. After normalizing the token-label probability matrix along a class dimension, the span of answers bounded by the begin and end token could be directly predicted.

\begin{table*}[!tbp]
    \centering\small
    \resizebox{1.0\linewidth}{!}{
    \begin{tabular}{l|c|c|cccccccc|c}
    \toprule

    \multirow{2}{*}{Method} & Total & Tunable & CoLA  & SST-2 & MPRC & QQP & $\text{MNLI}_\text{m}$ & $\text{MNLI}_\text{mm}$ & QNLI  & {RTE} & \multirow{2}{*}{AVG} \\

    {} & Params & Params & (8.5k) & (67k) & (3.7k) & (364k) & (393k) & (393k)  & (105k) & (2.5k) & {} \\

    \midrule
    % \multicolumn{12}{c}{\textit{Dev Set}} \\
    % \midrule
    $\text{BART}_\text{en}$-FT & 205M & \cellcolor{gray!50} 205M & 59.3 & 95.8 & 89.2 & 89.5 & 92.2 & 89.3 & 94.3 & 77.6 & 85.2 \\
    % \midrule
    $\text{BART}_\text{en}$-FbT & 223M & \cellcolor{gray!15}19M & 42.1 & 93.2 & 76.0 & \textbf{86.7} & 81.3 & 82.4 & \textbf{88.4} & 60.6 & 75.6 \\
    $\text{BART}_\text{en}$-$\mathcal{Y}$T & 220M & \cellcolor{gray!10}17M & \textbf{44.4} & \textbf{94.4} & \textbf{79.2} & 85.5 & \textbf{81.6} & \textbf{83.0} & 88.2 & \textbf{62.8} & \textbf{76.9} \\
    % \midrule
    % \multicolumn{12}{c}{\textit{Test Set}} \\
    \midrule
    $\text{BART}_\text{en}$-FT & 205M & \cellcolor{gray!50}205M & 51.4 & 95.6 & 86.4 & 73.4 & 89.4 & 88.6 & 94.5 & 73.9 & 80.6 \\
    % \midrule
    $\text{BART}_\text{en}$-FbT & 223M & \cellcolor{gray!15}19M & \textbf{41.7} & 90.3 & 74.0 & \textbf{65.3} & 81.8 & 81.4 & 88.0 & 56.6 & {71.1} \\
    $\text{BART}_\text{en}$-$\mathcal{Y}$T & 220M & \cellcolor{gray!10}17M & 40.9 & \textbf{95.6} & \textbf{76.8} & 64.2 & \textbf{82.5} & \textbf{82.4} & \textbf{88.1} & \textbf{57.4} & \textbf{72.2} \\
    \bottomrule
    \end{tabular}
    }
    \caption{Dev (the first block) and test (the second block) set results on GLUE benchmark using $\text{BART}_\text{LARGE}$'s encoder as PTM. Scores on test set are obtained from GLUE evaluation server. The metric for CoLA is Matthew’s Correlation while all other tasks use accuracy scores.}
    \label{tab:BART GLUE results}
\end{table*}

\section{Experimental Settings}
\subsection{Base Models}
We respectively use $\text{BART}_\text{LARGE}$\cite{lewis2019bart}, $\text{RoBERTa}_\text{LARGE}$\cite{liu2019roberta} and $\text{DeBERTa}_\text{XXL}$\cite{he2021deberta} as PTMs to evaluate our method since they contain representative PTMs for both NLG and NLU. We implement $\mathcal{Y}$-Tuning based on Hugging Face Transformers\cite{wolf-etal-2020-transformers}. One label representation is assigned for each class on downstream tasks. Similar to \citet{Lester2021Power}, those label embeddings are initialized from embeddings of tokens sampled from the vocabulary. The margin for triplet loss is set to 0.1 empirically.

\subsection{Baselines}
To verify the ability of $\mathcal{Y}$-tuning, we compare it to other tuning methods which also fix all the parameters of PTMs and do not need computing the gradients of PTMs during training.

\vspace{-0.5em}
\paragraph{Feature-based-Tuning (FbT)} Features generated by frozen PTM are directly fed into one tunable Transformer layer and then predicted by a classification head. We use this feature-based paradigm as our baseline.
\vspace{-0.5em}
\paragraph{$\mathcal{Y}$-Tuning ($\mathcal{Y}$T)} The PTM is fixed and the feature fuser is a Transformer layer. For BART we directly use its first layer of the decoder as our feature fuser. For encoder-only PTMs such as RoBERTa and DeBERTa, an additional layer added on top of the PTM is employed and $\mathcal{Y}$T is applied on the MNLI dataset to pretrain this decoder for other tasks. These settings are rather same as FbT except that the input embeddings are replaced with label representation and the linear head used in FbT is discarded. We further carry out experiments on $\mathcal{Y}$-tuning with decoder with different layer numbers. Suppose we have a decoder of $L_d$ layers, then the $i$th layer of the decoder would co-attend to the hidden states output by $\lfloor \frac{L_e}{i} \rfloor$th layers of encoder where $L_e$ denotes the number of encoder layers. We select $L_d$ from $\{1, 2, 4\}$ in our experiments. Borrowing from ALBERT's \cite{lan2020albert} approach, the parameters across layers are shared to ensure the number of tunable parameters remains the same. Moreover, the total FLOPs of the decoder in $\mathcal{Y}$-Tuning are $\mathcal{O}({L_d}NM)$, which is still much less than $\mathcal{O}(M^2)$ the baseline costs. We use $\mathcal{Y}\text{T}_n$ to denote that a $n$-layer weight-sharing decoder is used. In our experiments we use $\mathcal{Y}\text{T}_1$ on BART and $\mathcal{Y}\text{T}_4$ on RoBERTa and DeBERTa.

\vspace{-0.5em}
\paragraph{Fine-Tuning (FT)} All the parameters of the model can be updated. FT can be regarded as a measure of the upper limit of the model's performance.
\vspace{-0.5em}
\paragraph{Adapter-Tuning (AT)} The performance of Adapter-based Tuning is comparable with FT.\cite{houlsby2019parameter}. However AT saves tunable parameters only and it is time consuming compared to $\mathcal{Y}$T (Cf. Sec~\ref{sec:training efficiency}).
\vspace{-0.5em}
\paragraph{Prompt-Tuning (PT)} For prompt-tuning we report the results on SuperGLUE benchmark\cite{wang2020superglue} in accordance with \citet{Lester2021Power}. We just compare it on SuperGLUE benchmark since prompt-tuning is usually applied on few-shot learning.
\vspace{-0.5em}
\paragraph{WARP} WARP focuses on learning word embeddings concatenated to input texts. Their method is similar to PT while WARP takes an adversarial reprogramming approach. They also use MNLI to pretrain the added word embeddings.

\begin{table*}[!tbp]
    \centering\small
    \resizebox{1.0\linewidth}{!}{
    \begin{tabular}{l|c|c|c|c|ccccccc|c}
    \toprule

    \multirow{2}{*}{Method} & Total & Tunable & Training & Memory & \multirow{2}{*}{CoLA}  & \multirow{2}{*}{SST-2} & \multirow{2}{*}{MPRC} & \multirow{2}{*}{QQP} & \multirow{2}{*}{MNLI} & \multirow{2}{*}{QNLI}  & \multirow{2}{*}{RTE} & \multirow{2}{*}{AVG} \\
    {} & Params & Params & SpeedUp & Usage(\%) &  {} & {} & {} & {} & {} & {} & {} & {} \\

    \midrule
    $\text{RoBERTa}$-AT\(^\dagger\) & 355M & 3M &\cellcolor{gray!45} 0.6x &\cellcolor{gray!45} 88.7 & \bf67.4 & \bf96.3 & \bf92.9 & \bf88.5 & \bf90.4 & \bf94.7 & 83.4 & \bf87.7 \\
    $\text{RoBERTa}$-{WARP}\(^\dagger\) & 355M & $\leq\text{1M}$ & \cellcolor{gray!30}1.8x &\cellcolor{gray!30} 71.6 & 60.6 & 96.0 & 91.2 & 84.5 & 88.2 & 93.5 & 86.3 & 85.8 \\
    % $\text{RoBERTa}$-FbT & 368M & 14M & \multirow{2}{*}{\bf{3.2x}} & {} & 54.2 & 93.6 & 81.4 & 86.3 & 77.9 & 84.8 & 79.1 & 79.6 \\
    $\text{RoBERTa}$-$\mathcal{Y}\text{T}$ & 372M & 17M & \cellcolor{gray!15}{3.2x} &\cellcolor{gray!15} 18.1 & {54.4} & {94.5} & {85.0} & {87.4} & {83.1} & {88.2} & {81.9} & {82.1} \\
    $\text{DeBERTa}$-$\mathcal{Y}\text{T}$ & 1.6B & 31M & {\cellcolor{gray!30}1.6x} &\cellcolor{gray!15} 26.1 & {65.8} & {96.2} & {90.9} & 87.8 & {87.8} & {93.6} & \bf89.2 & {87.4} \\
    \midrule
    $\text{RoBERTa}$-FT\(^\dagger\) & 355M & 355M & \cellcolor{gray!60} 1x &\cellcolor{gray!60} 100 & 68.0 & 96.4 & 90.9 & 92.2 & 90.2 & 96.4 & 86.6 & 88.7 \\
    $\text{DeBERTa}$-FT\(^\ddagger\) & 1.6B & 1.6B & {-} & {-} & \bf72.0 & \bf97.2 & \bf93.1 & \bf92.7 & \bf91.8 & \bf96.0 & \bf93.5 & \bf90.9 \\
    % $\text{DeBERTa}$-FbT & 1.6B & 28M & \multirow{2}{*}{-} & \multirow{2}{*}{-} & 56.0 & 95.0 & 83.8 & \textbf{88.0} & 84.0 & 93.2 & \textbf{89.9} & 84.3 \\

    \bottomrule
    \end{tabular}
    }
    \caption{Dev set results on GLUE benchmark with $\text{RoBERTa}_\text{LARGE}$ and $\text{DeBERTa}_\text{XXL}$ backbone. Methods with \(^\dagger\) and \(^\ddagger\) indicate results reported in \citet{Hambardzumyan2021WARP} and \citet{he2021deberta} seperately. To profile training speedup and memory usage, we reuse frozen features to acclerate our method and details are discussed in Sec.~\ref{sec:training efficiency}. Here we denote $\text{RoBERTa}$-FT's training speedup as 1x and memory usage as 100\%. We show that $\text{DeBERTa}$-$\mathcal{Y}\text{T}$ achieves on-par performance with $\text{RoBERTa}$-FT and $\text{RoBERTa}$-AT but using less training time and saving much more memory.}  
    \label{tab:RoBERTa and DeBERTa GLUE results}
\end{table*}

\section{Results}

We measure the performance and amount of trainable parameters on GLUE benchmark and in Sec.~\ref{sec: quatitative results} quantitative results are represented. Sec.~\ref{sec:training efficiency} demonstrates advantages of our method in training-efficiency over other light-weight tuning methods. In addition, the model-robustness of $\mathcal{Y}$-Tuning is investigated in Sec.~\ref{sec:robustness} and promising results are exhibited. 
Besides, the generality of our method on sequence labeling and QA tasks is shown in Sec.~\ref{sec:results of generalized y-tuning}. More details are included in Appendix.

\subsection{Quantitative Results} \label{sec: quatitative results}
We select seven classification tasks from GLUE including CoLA, SST-2, MRPC, QQP, MNLI, QNLI and RTE to test $\mathcal{Y}$-Tuning's ability. For small datasets including CoLA, MRPC and RTE, we select a training batch size of 16. For other datasets, the training batch size is set to 32. We sweep learning rates in $\{1\cdot10^{-5}, 2\cdot10^{-5}, 3\cdot10^{-5}, 4\cdot10^{-5}\}$. Training epochs are 10 for all tasks. With different random seeds, we run each experiment 3 times and report the best result.

Results of BART model on development set and test set are demonstrated in Table \ref{tab:BART GLUE results}. $\mathcal{Y}$-Tuning improves the baseline by more than 1 point on both dev set and test set while requiring fewer tunable parameters and computation costs. Our method obtains gains on most datasets, especially for the SST-2 dataset where $\mathcal{Y}$-Tuning achieves comparable results with fine-tuning on the test set, approximately $5.3\%$ higher than the baseline. We speculate that this is because the categories of the SST-2 dataset can be explicitly described using labels (i.e. Great and Terrible). Therefore, the label perceiver could collect semantic information more straightforward.

Table \ref{tab:RoBERTa and DeBERTa GLUE results} shows results of RoBERTa and DeBERTa model on GLUE development set. Here we choose a 4-layer weight sharing decoder for better performance and the discussion about the number of decoder layers can be seen in Sec. \ref{sec:number of decoder layers}. The version of $\text{DeBERTa}_\text{XXL}$ is DeBERTa-V2 finetuned with MNLI task, which slightly improves the results of other tasks. We choose this PTM for $\text{DeBERTa}$-$\mathcal{Y}\text{T}$ on all tasks except for MNLI where the original DeBERTa-V2 model without MNLI finetuning is used. 

Although it's impossible to fine-tune all parameters of $\text{DeBERTa}_\text{XXL}$ with 1.6 billion parameters on a single RTX 3090 GPU due to out of memory(OOM) issue, $\mathcal{Y}$-Tuning works well and achieves performance more than $96\%$ of full fine-tuning with only $2\%$ tunable parameters, hence requires quite small storage space per task. 
The results of $\text{DeBERTa}$-$\mathcal{Y}\text{T}$ is competitve with $\text{RoBERTa}$-FT but $\text{DeBERTa}$-$\mathcal{Y}\text{T}$ is even faster and more memory usage economize than $\text{RoBERTa}$-FT during training with feature reusing method described in Sec.\ref{sec:training efficiency}. From the table we can also see that $\text{DeBERTa}$-$\mathcal{Y}\text{T}$ is more than 3 times faster $\text{RoBERTa}$-AT and only uses 30\% memory that of $\text{RoBERTa}$-AT while achieving similar results.
It is noteworthy that as the number of PTM's parameters increases, the gap between fine-tuning and $\mathcal{Y}$-Tuning is narrowing. This means that our approach is expected to achieve comparable results to fine-tuning in the future, given the current trend of growing larger PTMs. 

We further validate our method on several tasks on SuperGLUE benchmark including RTE, BoolQ and CB in consistence with \citet{liu2021ptuning}. For these datasets we select epochs as 20 and sweeps learning rate from
$\{1\cdot10^{-5}, 2\cdot10^{-5}, 3\cdot10^{-5}, 1\cdot10^{-4}\}$. Table \ref{tab:BART SuperGLUE results} suggests that $\mathcal{Y}$-Tuning yields strong performance increase over both feature-based tuning and prompt-tuning\cite{Lester2021Power,liu2021gpt}.

\begin{table}[tb!]
    \centering
    \resizebox{1.0\linewidth}{!}{
    \begin{tabular}{l|c|c|ccc}
    \toprule
    \multirow{2}{*}{Method} & Total & Tunable & RTE & BoolQ & CB\\
    {} & Params & Params & (2.5k) & (9.4k) & (0.25k) \\
    \midrule

    $\text{RoBERTa}$-FT\(^\dagger\) & 355M & 355M & 86.6 & 86.9 & 98.2 \\
    $\text{RoBERTa}$-PT\(^\dagger\) & 355M & - & 58.8 & 62.3 & 71.4 \\
    \midrule
     $\text{RoBERTa}$-FbT & 368M & 14M & 78.3 & 70.9 & 89.3 \\
    $\text{RoBERTa}$-$\mathcal{Y}\text{T}$ & 372M & 17M & \textbf{82.7} & \textbf{75.2} & \textbf{92.3} \\

    \bottomrule
    \end{tabular}
    }
    \caption{Results on SuperGLUE dev set. Our method significantly surpasses feature-based baseline and prompt-tuning. Methods with \(^\dagger\) indicate results reported in \citet{liu2021ptuning}.}
    \label{tab:BART SuperGLUE results}
\end{table}

\subsection{Training Efficiency}\label{sec:training efficiency}
We compare training speed of fine-tuning, adapter-based tuning~\cite{pfeiffer2021adapterfusion}, prompt tuning~\cite{liu2021gpt}, WARP\cite{Hambardzumyan2021WARP} and $\mathcal{Y}$-Tuning. Adapter-based tuning method in our experiments is implemented with their open source code\footnote{\url{https://github.com/Adapter-Hub/adapter-transformers}}~\cite{pfeiffer2020adapterhub}. Prompt tuning results are reproduced with code\footnote{\url{https://github.com/THUDM/P-tuning-v2}} released by \citet{liu2021ptuning} with prompt length as 8. WARP results are obtained with code\footnote{\url{https://github.com/YerevaNN/WARP}} provided by \citet{Hambardzumyan2021WARP}. We use $\mathcal{Y}\text{T}_1$ as our competing method. All these systems  are  evaluated  under with same experimental environments and the same batch size with $\text{RoBERTa}_\text{LARGE}$ as PTM. Concretely, we run these tests on a single 24G GeForce RTX 3090 GPU for 3 times and report the average results.

\begin{figure}[t]
    \centering
    \resizebox{1.0\linewidth}{!}{
    \includegraphics[width=1.0\linewidth]{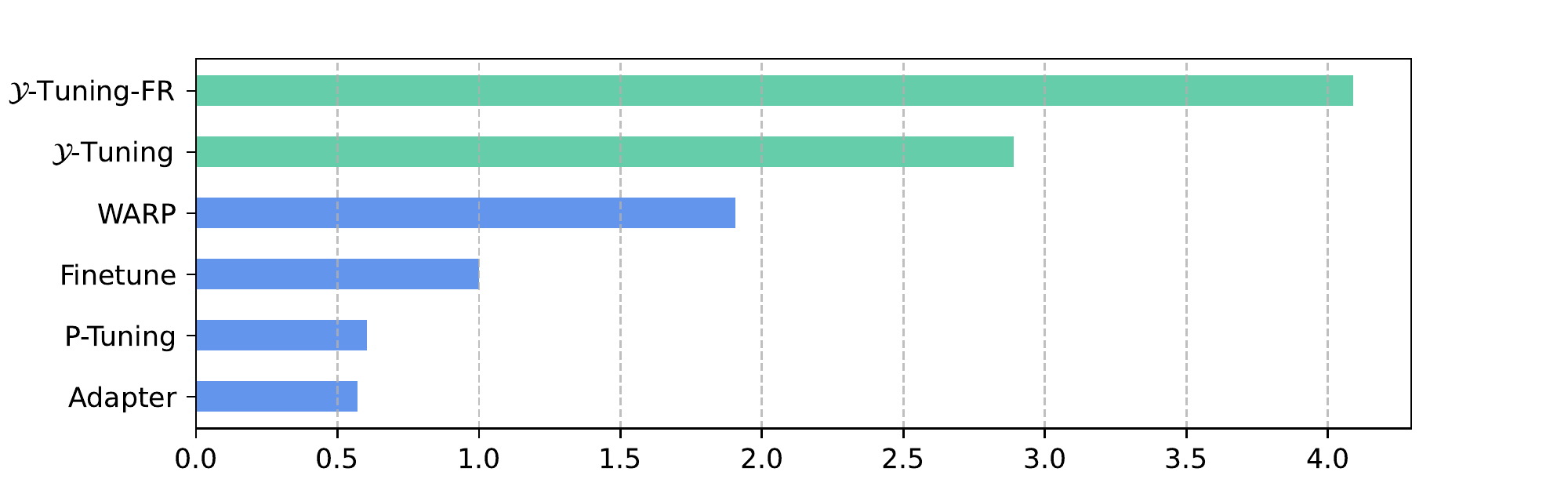}
    }
    \caption{Training speed-up ratio over fine-tuning on various methods.}
    \label{fig:speed_up_ratio}
\end{figure}

The comparison results are shown in Figure \ref{fig:speed_up_ratio}. $\mathcal{Y}$-Tuning is 2.8 times faster than directly fine-tuning. It should be noted that for $\mathcal{Y}$-Tuning we can save features representations of  PTM  on disks during the first epoch and reuse them for subsequent epochs, which further reduces the overhead of computation tremendously (denoted as $\mathcal{Y}$-Tuning-FR in Figure \ref{fig:speed_up_ratio}). Since storage costs are much cheaper than computing costs in industry, our approach is of practical value for saving training costs.

On one hand, our method does not cost a lot of GPU memory. $\mathcal{Y}$-Tuning consumes $10\%\sim20\%$ memory that of fine-tuning, while WARP requires more than $70\%$, adapter-based method more than $80\%$ and P-Tuning more than $90\%$ GPU memory compared with that of fine-tuning. This means that our method can be well applied to training scenarios with limited device memory resources compared with other lightweight tuning methods. On the other hand, we also obtain an obvious drop in GPU usage. The GPU usage with $\mathcal{Y}$-Tuning-FR is only $20\%$ that of fine-tuning, which indicates that our approach has great advantages while simultaneously training for different tasks.

\begin{figure}[h]
    \centering
    \hspace{-1.5em}
    \subfloat[SST-2 test set]{\label{fig:SST-2_robust}
      \includegraphics[width=0.5\linewidth]{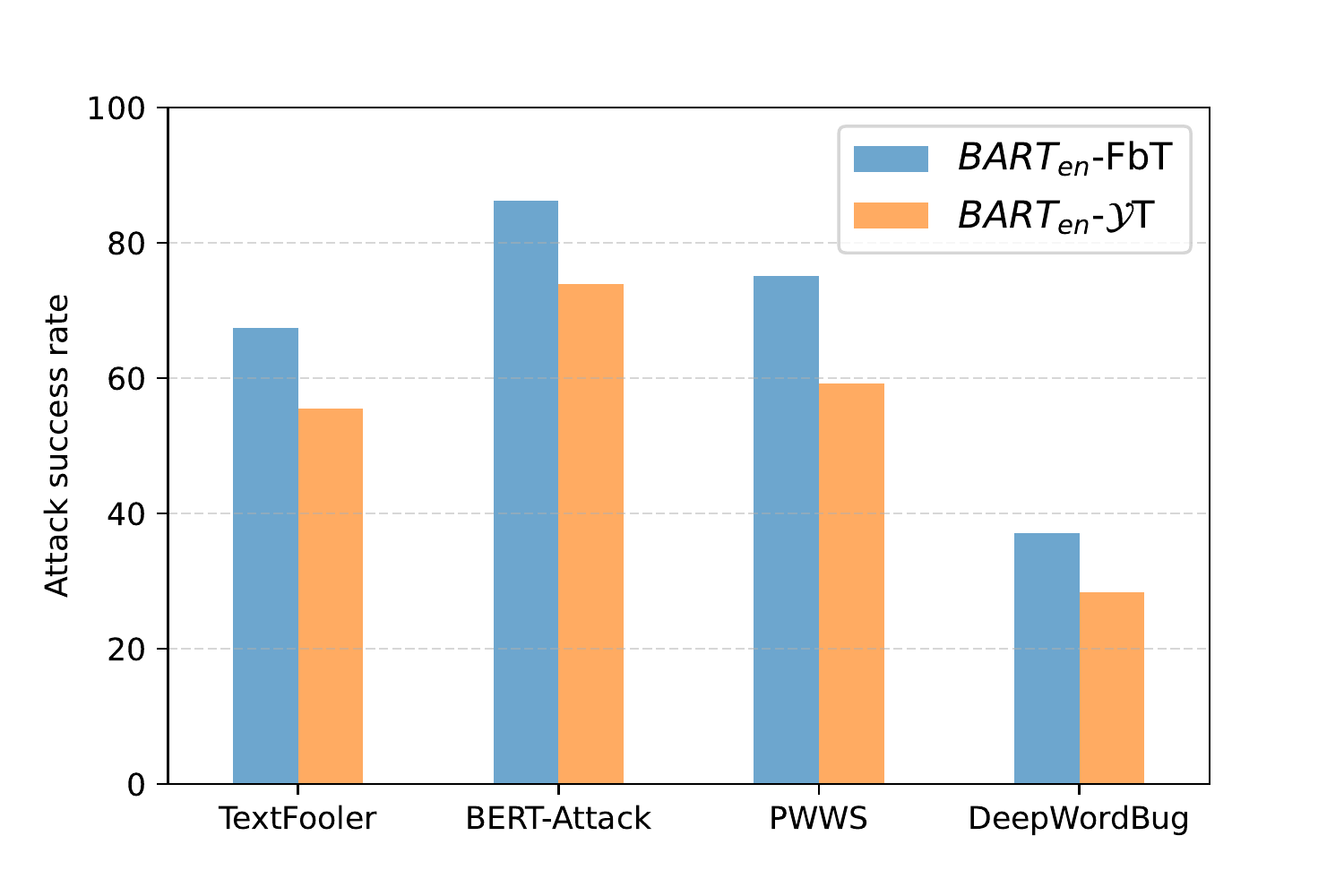}
    }
    %\newline
    \hspace{-1.5em}
    \subfloat[$\text{MNLI}_\text{m}$ test set]{\label{fig:MNLI-m_robust}
      \includegraphics[width=0.5\linewidth]{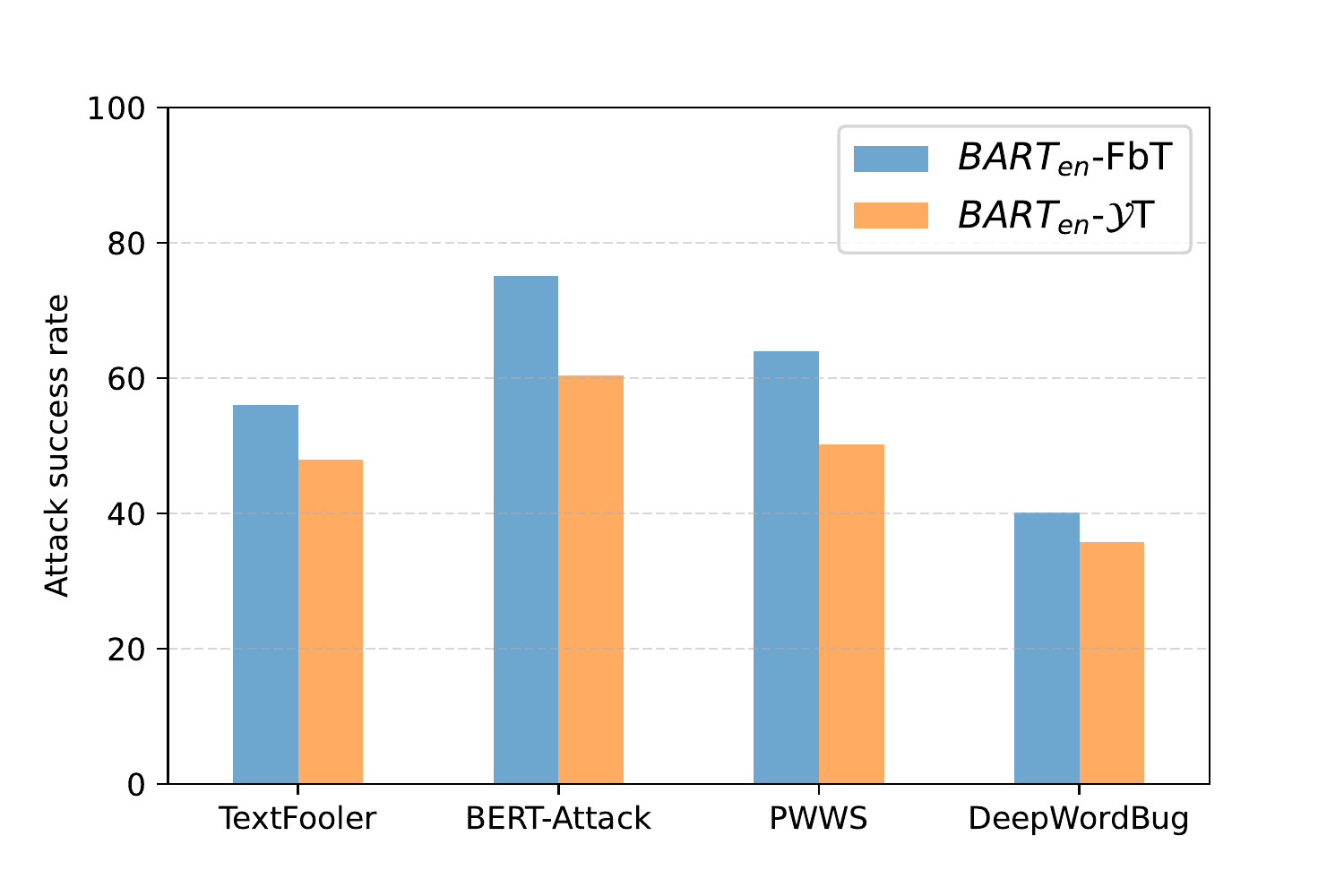}
    }
    \caption{Robustness against Adversarial Attacks.}
    \label{fig:Robustness Analysis}
\end{figure}

\subsection{Robustness} \label{sec:robustness}

Compared with the simple mapping of complex sentence features to low-dimensional space, $\mathcal{Y}$-Tuning uses an equally complex label representation to match the semantic representation of sentences, thus alleviating the problem of robustness deterioration caused by dimension reduction in the mapping process. We validate the robustness of our methods by attacking our models on SST-2 and MNLI datasets with different adversarial attackers. The baseline and $\mathcal{Y}$-Tuning are attacked by four typical attack models covering both word-level and char-level types including TextFooler~\cite{jin2019bert}, BERT-Attack~\cite{li2020bertattack}, PWWS~\cite{ren2019generating} and DeepWordBug~\cite{gao2018black} implemented in \citet{zeng2020openattack}. We evaluate both methods on the full SST-2 test set and 1k samples randomly selected from MNLI-matched test set using attack successful rate (ASR) as the metric. ASR represents the rate of model output categories changing after attacker methods are applied. The lower ASR is, the more robust the victim model is. As seen in Figure \ref{fig:Robustness Analysis}, the ASR of $\mathcal{Y}$-Tuning is $5\%\sim15\%$ lower than the feature-based baseline, which shows that our method is more insensitive to perturbation.

\subsection{Results of Generalized $\mathcal{Y}$-tuning} \label{sec:results of generalized y-tuning}
\begin{table}[t]
    \centering
    \resizebox{1.0\linewidth}{!}{
    \begin{tabular}{l|cc|c}
    \toprule
    \multirow{2}{*}{Method} & CoNLL03 & CoNLL03 &  \multirow{2}{*}{SQuAD 1.0} \\
    {} & NER & CHUNK & {} \\
    \midrule

    $\text{BART}$-FT & 95.6 & 91.8 & 92.0 \\
    $\text{RoBERTa}$-PT\(^\dagger\) & 86.1 & - & 12.0 \\
    \midrule
     $\text{BART}$-FbT & 70.9 & 73.6 & 73.6 \\
    $\text{BART}$-$\mathcal{Y}\text{T}$ & \textbf{88.2} & \textbf{85.9} & \textbf{82.7} \\

    \bottomrule
    \end{tabular}
    }
    \caption{Generalized $\mathcal{Y}$-Tuning on sequence labeling and span-based QA tasks. F1 score is used as metric across tasks. Methods with \(^\dagger\) indicate results reported in \citet{liu2021ptuning}.}
    \label{tab:Generalized Y-Tuning}
\end{table}
We adopt CoNLL03 NER and CHUNK datasets to evaluate the performance of $\mathcal{Y}\text{T}$ on sequence labeling tasks. For span-based question answering task we use SQuAD 1.0 dataset. As shown in Table~\ref{tab:Generalized Y-Tuning}, Generalized $\mathcal{Y}$-tuning presents far superior performance than FbT on sequence labeling and span-based QA tasks. Class imbalance issue is quite regular for sequence tagging and extractive QA tasks: negative examples noted as tag \texttt{"O"} 
dominate training and lead to poor generalization ability~\cite{li-etal-2020-dice}. $\mathcal{Y}$-Tuning provides such a solution to learn semantic label representations, and hence alleviates the class imbalance issue. Prompt-Tuning shows poor performance on QA even with a stronger PTM and ~\citet{liu2021ptuning} suggests that this is because the task is too challenging.

\section{Related Work}

$\mathcal{Y}$-tuning is related to two lines of work: label embeddings and text matching.

\paragraph{Label Embedding}
Label embedding~\cite{yeh2017learning,sun2017label,wang2020incorporating} is to enhance feature representation by integrating label information, which is an effective strategy if the output space is complex and correlative.
\citet{Hambardzumyan2021WARP} improve prompt-tuning by initializing
the weights of the output layer with the
word embeddings used for the input, which can also be regarded as label embedding.

Different from label embedding, $\mathcal{Y}$-tuning builds a more complex label representation and inject it with the label-specific feature. $\mathcal{Y}$-tuning can be regarded as a wrapper of PTMs, rather than just label embeddings.

\paragraph{Text Matching} Text matching~\cite{sun-etal-2019-utilizing,Chai20Description,Wang21Entailment} is also a framework to reformulate classification problem, in which
we can predict whether the pair-wise input $(x, y_c)$ is matched, where $x$ is the original text and $y_c$ is the natural language description of label $c$. Existing methods usually concatenate $x$ and $y_c$ into a single sequence $x\oplus y_c$, use PTMs to predict their score. According to the input mode of PTMs, text matching is more like prompt-tuning. Different from text matching, $\mathcal{Y}$-tuning is more like a wrapper of PTMs.

\section{Conclusion}

In this paper, we  explore a new tuning paradigm, $\mathcal{Y}$-tuning, to condition  frozen pre-trained models to perform specific downstream tasks. Instead of tuning the complicated feature representation into simple label space, $\mathcal{Y}$-tuning aims to build a more powerful label space and adapt it to feature space. With this tuning paradigm, we can train PTMs efficiently with less time and memory consuming than other tuning paradigms, which enables tuning very-large-scale models such as GPT-3 possible. $\mathcal{Y}$-tuning also saves storage space a lot because it requires very few tunable parameters. Moreover, $\mathcal{Y}$-tuning shows generally applicability on various NLU tasks.

\section*{Limitations}
Although we have obtained some promising results, there are still much room for improvement. There is still a performance gap between fine-tuning and $\mathcal{Y}$-tuning and more effective architecture of $\mathcal{Y}$-tuning still needs further exploration. And for encoder-only PTMs like RoBERTa, we will need to pretrain the feature-fuser if downstream training set is small. Otherwise the parameters of Transformer-based decoder are initialized randomly and might deteriorate performance.

%\section*{Acknowledgements}

% \bibliography{anthology,acl}
% \bibliographystyle{acl_natbib}

\clearpage
\appendix

\section*{Appendix}

\section{Ablation Experiments}
\label{sec:ablations}

We perform an ablation to study how much the number of embeddings corresponding to each class, the initial label embeddings and the number of decoder layers would have influence on $\mathcal{Y}$-Tuning. Experimental results show that our method is insensitive to these settings.

\subsection{Affects on the Number of Label Embeddings per Label}
\label{sec:exp-emb-num}

\begin{figure}[!ht]
    \centering
    \resizebox{0.9\linewidth}{!}{
    \includegraphics[width=1.0\linewidth]{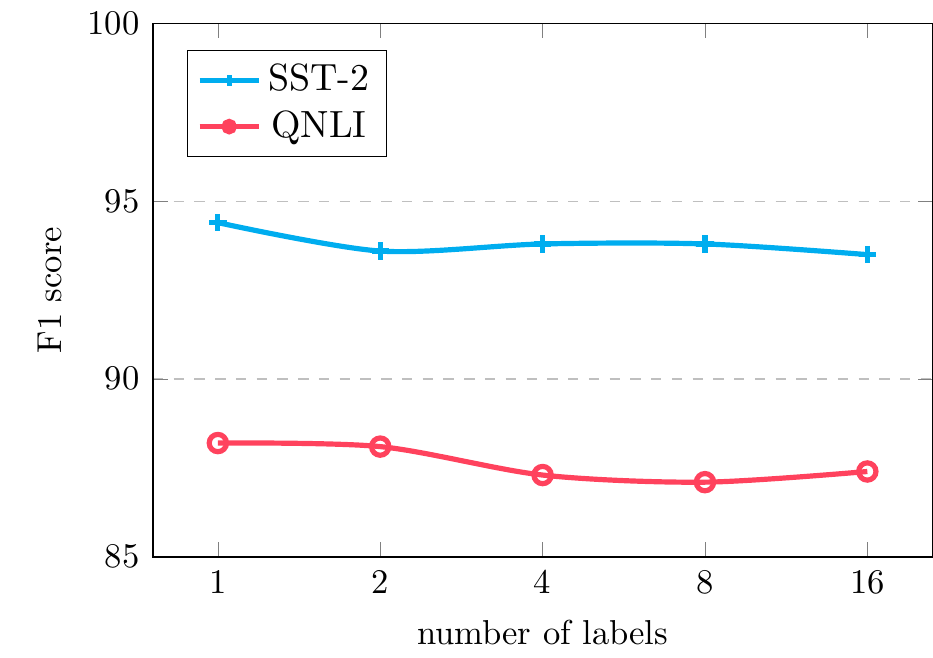}
    }
    \caption{Ablation of the number of label embeddings per label. }
    \label{fig:ablation_label-num}
\end{figure}

First we conduct experiments on the number of label embeddings. We assign different number of labels for each category and calculate margin loss for each pair of positive and negative labels during training. The summation of logits belonging to each class represents the final score. Figure \ref{fig:ablation_label-num} suggests one label embeddings is sufficient for achieving best results. 

\subsection{Affects on Initialization of  Label Embeddings}
\label{sec:exp-init}

We instigate whether the label embeddings initialization matters. In consistence with \citet{Lester2021Power}, we ablate the initialization in these ways:
\begin{enumerate} [a)]
    \item {\textbf{Random Uniform.}} Uniformly distributed randomization in a range of [-0.5, 0.5].
    \item {\textbf{Sampled Vocab.}} Embeddings of tokens sampled from vocabulary in the order of descending frequency of pre-training.
    \item {\textbf{Class Label.}} Embeddings of class labels. (e.g. `great' for positive sample and `terrible' for negative sample on SST-2). We average the embeddings if the class label is split into multiple subwords.
    \item {\textbf{Opposite Label.}} Embeddings of opposite class labels (e.g. `terrible' for positive sample and `great' for negative sample on SST-2).
\end{enumerate}

\begin{table}[!h]
    \centering
    \resizebox{0.6\linewidth}{!}{
    \begin{tabular}{l|c}
    \toprule
    Initialization & SST-2 \\
    \midrule
    Random Uniform & 93.8 \\
    Sampled Vocab & \textbf{94.4} \\
    Class Label & 94.2 \\
    Opposite Label & 93.8 \\
    \bottomrule
    \end{tabular}
    }
    \caption{Ablation of label embeddings initialization.}
    \label{tab:initialization abaltion}
\end{table}

Besides, the task token is initialized as `<s>'. As shown in Table \ref{tab:initialization abaltion}, while sampled from vocab slightly outperforms other initialization manners, assigning opposite label embeddings still achieves superior results over baseline. This indicates that our method works well for various label embeddings initialization, which means that our method is also suitable for those classification tasks without explicit tokens corresponding to class labels.

\begin{table}[htbp!]
    \centering
    \resizebox{1.0\linewidth}{!}{
    \begin{tabular}{l|c|c|cc}
    \toprule
    \multirow{2}{*}{Method} & Total & Tunable & \multirow{2}{*}{SST-2} & \multirow{2}{*}{$\text{MNLI}_\text{m / mm}$} \\

    {} & Params & Params & {} & {} \\

    \midrule
    $\text{RoBERTa}$-FT & 355M & 355M & 96.4 & 90.4 / 90.1 \\
    \midrule
    $\text{RoBERTa}$-FbT & 368M & 14M &  92.4 & 77.4 / 78.4 \\
    $\text{RoBERTa}$-$\mathcal{Y}\text{T}_1$ & 372M & 17M & 92.5 & 76.4 / 77.2 \\
    $\text{RoBERTa}$-$\mathcal{Y}\text{T}_2$ & 372M & 17M & 93.8 & 80.7 / 81.0 \\
    $\text{RoBERTa}$-$\mathcal{Y}\text{T}_4$ & 372M & 17M & \textbf{94.5} & \textbf{82.8} / \textbf{83.3} \\
    \bottomrule
    \end{tabular}
    }
    \caption{Evaluation results on GLUE benchmark with $\text{RoBERTa}_\text{LARGE}$ backbone with different number of decoder layers.}
    \label{tab:RoBERTa Layer Ablation}
\end{table}

\subsection{Number of Decoder Layers}
\label{sec:number of decoder layers}
Table \ref{tab:RoBERTa Layer Ablation} shows experimental results of $\mathcal{Y}$-Tuning on $\text{RoBERTa}_\text{LARGE}$ with different number of decoder layers with parameters initialized randomly. We find that the performance of using only a one-layer decoder was not very ideal in this setting. We suppose that the lack of low-level semantic details would result in the difficulty of optimization. However, as the number of decoder layers increases, our method improves significantly and surpasses our baseline by more than $2\%$ on SST-2 and $6\%$ on MNLI with fewer FLOPs and comparable parameters. Moreover, $\mathcal{Y}\text{T}_4$ only requires an additional $7\%\sim10\%$ time consumption than $\mathcal{Y}\text{T}_1$ during both training and inference in our experiments, which is quite marginal considering the bringing performance boost.

\end{document}